\documentclass[10pt]{article}

\usepackage[utf8]{inputenc}
\usepackage[T1]{fontenc}
\usepackage{times}
\usepackage[margin=1.5in]{geometry}
\usepackage{hyperref}
\usepackage{url}
\usepackage{booktabs}
\usepackage{amsfonts}
\usepackage{nicefrac}
\usepackage{microtype}
\usepackage{xcolor}
\usepackage{graphicx}
\usepackage{natbib}
\usepackage{tabularx}
\usepackage{array}
\usepackage{float}
\usepackage{tikz}
\usepackage{ragged2e}
\usepackage{caption}
\usetikzlibrary{arrows.meta,positioning,calc}
\usepackage{titlesec}
\titleformat{\section}{\large\bfseries}{\thesection}{1em}{}
\titleformat{\subsection}{\normalsize\bfseries}{\thesubsection}{1em}{}

\title{\textbf{The Evaluation Trap: Benchmark Design as\\ Theoretical Commitment}}

\author{
  Theodore J. Kalaitzidis \\
  Brown University \\
  \texttt{tj.k@brown.edu}
}

\date{}

\begin{document}

\maketitle

\begin{abstract}
Every AI benchmark operationalizes theoretical assumptions about the capability it claims to assess. When assumptions function as unexamined commitments, benchmarks stabilize the dominant paradigm by narrowing what counts as progress. Over time, narrow evaluation reorganizes capability concepts: architectures and definitions are selected for benchmark legibility until evaluation ceases to track an independent object and instead produces a version of the target defined by its own operational assumptions. The result is a trap: evaluation frameworks treat self-reinforcing assessments as valid, both creating and obscuring structural limits on what the current paradigm can accomplish. We introduce \textit{Epistematics}, a methodology for deriving evaluation criteria directly from technical capability claims and auditing whether proposed benchmarks can discriminate the claimed capability from proxy behaviors. The contribution is meta-evaluative: an audit procedure, a failure mode taxonomy, and benchmark-design criteria for evaluating capability-evaluation coherence. We demonstrate the procedure through a worked audit of \citet{dupoux2026}, a proposal that revises the dominant paradigm's theoretical assumptions at the architectural level while reproducing them in its evaluation criteria, thereby entrenching the constraint it seeks to overcome in a form the evaluation cannot detect.
\end{abstract}

\section{Introduction}

It is well established that evaluation does not neutrally measure performance; it shapes research and development \citep{raji2021}, contaminating the object of study. What remains underexplained are the mechanisms by which this occurs and how it might be addressed.

This paper investigates evaluation within AI, arguing that benchmarks shape the field because they operationalize implicit theoretical assumptions that govern how capabilities are modeled, evaluated, and ultimately judged valid. As benchmark success becomes the dominant signal of progress, it becomes the de facto interpretation of capability, obscuring the distinction between performance and the capacities it is taken to represent. This dynamic blinds the discipline to its own boundaries, causing it to treat structural dead-ends as mere technical hurdles. We name this contamination the \emph{evaluation trap} and propose a methodological response. This dynamic is continuous with the epistemic laundering mechanisms described in \citet{kalaitzidis2026}, in which contested theoretical commitments are stabilized and naturalized through AI discourse.

We introduce Epistematics as a meta-evaluative framework for addressing this problem. The methodology proceeds in four steps: (1) specify the claimed capability; (2) extract the theoretical assumptions that define what it requires; (3) derive the architectural and environmental requirements those assumptions imply; and (4) test whether the proposed evaluation criteria can discriminate the claimed capability from simpler proxy behaviors. The output is tractable design input: discriminative evaluation criteria, predicted failure modes, and candidate benchmark specifications. Existing benchmark analysis typically identifies evaluation failure after the fact; Epistematics derives evaluative conditions from the capability claim itself, prior to benchmark construction.

We demonstrate the procedure through a worked audit of \citet{dupoux2026}, whose proposal for ``autonomous learning'' attempts to revise the dominant AI paradigm's theoretical assumptions at the architectural level while reproducing them in its evaluation criteria. Through this demonstration, we make three contributions to evaluation science: a formalized audit procedure for diagnosing capability-evaluation incoherence; a worked analysis demonstrating how evaluation failure persists even within state-of-the-art research; and design criteria for constructing benchmarks capable of detecting the capabilities they claim to measure.

\section{The mechanisms of the evaluation trap: transferability, circularity, and behavioral approximation}

To assess any capability, a benchmark must specify what counts as evidence, what performances are diagnostic, and what conditions would reveal its absence. These specifications encode theoretical assumptions about the target; once operationalized, they function as theoretical commitments that determine what the framework can and cannot detect \citep{bowker1999}. Evaluation that does not examine its own assumptions cannot detect the structural limits of its own methods. A field that relies on such evaluation cannot generate the revisions required to surpass those limits. This dynamic operates through three interdependent mechanisms: the \textit{transferability assumption} that converts benchmark performance into capability attribution, the \textit{circularity} through which evaluation progressively reorganizes its own object, and the \textit{behavioral approximation} that results.

\subsection{The transferability assumption}

In AI, theoretical assumptions permeate the design stack. Objective functions determine optimization targets, training data delimit relevant evidence, evaluation conditions define valid tests, and metrics specify success. Together, these choices instantiate a specific theory of the target capacity and foreclose alternative theories. For example, a benchmark that rewards output similarity to a target distribution does not measure the capability in general; it measures the capability \textit{as defined by a distributional performance theory} and motivates the construction of architectures to realize it.

A benchmark that operationalizes such a \textit{de facto} theory cannot distinguish between a system that improves on its own criteria and one that instantiates the capacity it claims to measure \citep{chollet2019,raji2021}. This limitation is structural: making the distinction would require a comparative theory the framework does not encode.

Inference from benchmark performance to capability attribution depends on an assumption of \emph{transferability}: that performance on a benchmark licenses inference to the broader capability \citep{cartwright1999}. Without this theoretical assumption, the gap between what the benchmark measures and what the capability requires would be visible and addressable. With it, observable performance is treated as evidence of capability, even when the mechanisms that would justify that inference are absent. Under optimization pressure, this inference reshapes the original target. Once performance is treated as sufficient evidence of capability, optimization operates on the benchmark's operationalization rather than the capability itself. This establishes a reinforcing feedback loop: the transferability assumption licenses performance as capability, optimization follows, and the target reorganizes around what the benchmark can detect.

\subsection{The circularity problem}

We call this loop the \emph{circularity problem}. It differs from data contamination, which describes test-set leakage from training data, and extends beyond Goodhart's Law, which assumes a stable target that proxy measures fail to track \citep{goodhart1984}. While Goodhart's Law identifies the divergence between a stable target and an increasingly manipulated proxy, circularity identifies a feedback loop in which the act of evaluation participates in the conceptual and technical constitution of the target itself. This is a ``looping effect'' \citep{hacking1999} in which classifications participate in bringing into being the phenomena they describe. As evaluation frameworks stabilize, they reshape what counts as the capability \citep{porter1995}, producing systems that satisfy evaluation criteria while stabilizing those criteria as the capability itself. This stabilization process operates through mechanisms analyzed in \citet{kalaitzidis2026} as epistemic laundering: the naturalization of misrecognition through double concealment.

This produces a ``lossy compression'' of capability concepts in which the diverse theoretical commitments required to instantiate a capability are discarded while tractable benchmark-legible proxies are retained. This reduction generates a \emph{structural ceiling}: a limit on what a paradigm can achieve or recognize as achievement. Because the underlying assumptions fix (1) the problem space, (2) the space of legible solutions, and (3) the criteria for evidence of progress, the ceiling is not visible from within the paradigm; it appears not as a boundary but as the horizon of the possible \citep{kuhn1962}.

\subsection{The lossiness of behavioral approximation}

Structural ceilings create blind spots in the field: some capabilities sought by frontier research may require theoretical commitments that are incommensurable with those assumed by dominant benchmarks. The incompatibility is expressed at the level of architecture. For example, a distributional theory is typically operationalized through fixed-weight deployed models, while a cybernetic theory\footnote{Cybernetics, developed by Wiener [1948], treats learning as real-time error correction through feedback loops coupling system outputs to environmental states.} requires closed feedback loops coupling system outputs to environmental states. Optimization within the distributional evaluation framework not only fails to approach cybernetic capabilities; it creates the appearance that existing approaches are sufficient with enough scale \citep{chollet2019,marcus2020,agre1997}. Evaluation and engineering become mutually reinforcing, obscuring the alternative theoretical frameworks, architectures, and evaluations that might support them.

This reinforcing circularity produces a specific class of systems. Current benchmarks evaluate defined problem spaces with stable conditions and verifiable outputs; many capabilities AI engineers seek (e.g., autonomous learning, genuine reasoning, adaptive world-modeling) operate under open-ended, context-dependent conditions and resist predefined success criteria. Benchmarks measure these adaptive capabilities by converting ill-structured problems into well-structured proxies \citep{jonassen1997}, and systems optimized against the proxies learn to succeed under the conditions the proxies define \citep{chollet2019,geirhos2020}. We call the resulting systems \emph{behaviorally approximate}: they produce outputs that resemble those of a genuinely capable system under benchmark conditions without instantiating the mechanisms those outputs require.

Behavioral approximation arises from benchmark design under the transferability assumption and from optimization within a circular, lossy evaluation regime. It is the form in which structural ceilings manifest. Performance fails to transfer not because the system failed to learn, but because it learned precisely what the benchmark made legible.

This categorical oversight evades the evaluation framework that produces it. When failures occur, practitioners address them through the technical rationality of engineering practice \citep{schon1983}: more data, architectural refinement, extended training. But if the ceiling is structural rather than technical, these responses cannot resolve it. They further optimize within the same assumptions that produced the limitation, which is why the field's most ambitious capabilities may remain out of reach despite continued progress within dominant evaluation regimes. The field is caught in an evaluation trap: its evaluative frameworks cannot detect the gap between claimed and instantiated capability, and the only solutions available are those the paradigm can generate.

\section{Epistematics: a meta-evaluative procedure}

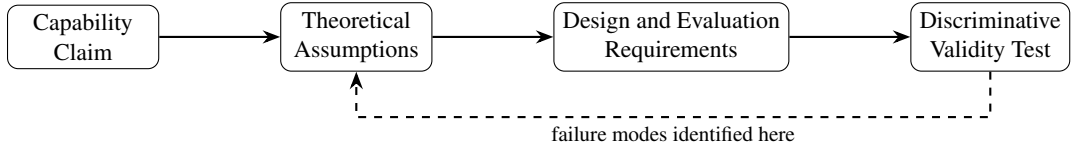
\begin{figure}[h]
\centering
\begin{tikzpicture}[
  node distance=1.6cm,
  box/.style={rectangle, draw, rounded corners, minimum width=2.0cm, minimum height=0.8cm, align=center, font=\small},
  arrow/.style={-{Stealth}, thick}
]
  \node[box] (A) {Capability\\Claim};
  \node[box, right=of A] (B) {Theoretical\\Assumptions};
  \node[box, right=of B] (C) {Design and Evaluation\\Requirements};
  \node[box, right=of C] (D) {Discriminative\\Validity Test};

  \draw[arrow] (A) -- (B);
  \draw[arrow] (B) -- (C);
  \draw[arrow] (C) -- (D);

  \draw[arrow, dashed] (D.south) -- ++(0,-0.6) -| node[below, font=\footnotesize, pos=0.25] {failure modes identified here} (B.south);
\end{tikzpicture}
\caption{The Epistematics procedure: a four-step process from capability claim to discriminative validity test, with a feedback arrow identifying failure modes.}
\end{figure}

Epistematics treats capability definitions as theoretical commitments that must be specified before evaluation. It proceeds in four steps: (1) specify the capability claim as used in technical and public discourse; (2) extract the theoretical assumptions the claim invokes or implies; (3) derive the architectural and environmental requirements necessitated by those assumptions; and (4) test whether proposed evaluation criteria can discriminate the target capability from proxy behaviors. A benchmark fails the discriminative validity test if a system lacking the target mechanism can satisfy it, or if a system possessing the target mechanism cannot be uniquely identified by it. Discriminative validity therefore concerns not whether a benchmark predicts performance, but whether it distinguishes the target mechanism from alternative routes to benchmark success.

The procedure is designed to be reproducible: independent researchers applying the four steps should converge on comparable architectural requirements, with disagreements traceable to differences in theoretical interpretation or constraint derivation. Such disagreement is itself a finding rather than a defect. The output is tractable design input: discriminative evaluation criteria, predicted failure modes, and candidate benchmark specifications that engineers can evaluate, contest, and develop. These outputs specify the conditions under which a benchmark can validly attribute a claimed capability.

Specifically, the failure modes of the discriminative validity tests are not generic benchmark flaws; they are specific signatures of the circularity problem. Each names a way the assumption of transferability sustains structural blindness in evaluation design.

\textbf{Proxy substitution.} The benchmark measures task convergence rather than the targeted capability. A system that achieves criterion performance by exploiting statistical regularities in the proxy task is indistinguishable from one that genuinely instantiates the target mechanism.

\textbf{Architectural indistinguishability.} Some mechanisms are structurally impossible within certain architectures regardless of training history. A fixed-weight, non-recurrent system cannot instantiate real-time feedback coupling, but trials-to-criterion measures outcome rather than mechanism and cannot register the structural absence. A system approximating feedback-driven behavior and one instantiating it become evaluatively equivalent.

\textbf{Context blindness.} A benchmark that does not vary context cannot detect whether competence is context-sensitive or task-specific. Performance on a fixed task family establishes only that the benchmark's own task was solved; the conditions under which general competence would distinguish itself from task-specific optimization never arise.

\textbf{Criterion leakage.} Statistical parity does not imply mechanistic equivalence. A system achieving human-level metrics (e.g., learning speed) through brute-force distributional optimization is indistinguishable from one achieving it through genuine adaptive coupling, because the criterion registers only the output and not the mechanism that produced it, allowing alternative mechanisms to pass undetected.

\textbf{Approximation ceiling.} A system optimized against a well-structured proxy learns to satisfy benchmark criteria under the conditions the proxy defines. Because those conditions are within the system's design envelope, the benchmark cannot detect failure at the point where the ill-structured capability the proxy represents would actually be required. A system exhibiting genuine adaptive capacity and one exhibiting behavioral approximation are indistinguishable within the benchmark's defined conditions; they diverge only outside them, which the benchmark cannot reach.

\subsection{Falsifiability and generalizability}

If Epistematics-derived criteria fail to distinguish systems with the target mechanism from proxy-performing systems, or fail to improve discriminative validity over standard benchmarks, the framework should be rejected. Empirical falsification would require demonstrating that standard benchmarks predict capability-relevant behavior in genuinely novel conditions at least as well as Epistematics-derived criteria; that is, that mechanism-level discrimination offers no predictive advantage over performance-level evaluation.

The procedure is also domain-general: the four steps apply to any capability claim whose theoretical sources yield derivable architectural requirements. Applied to reasoning, Epistematics requires distinguishing inferential operations that preserve validity across novel compositional structures from pattern completion within the training distribution \citep{chollet2019,marcus2020}. Applied to world-modeling, it requires distinguishing representations updated through prediction error against environmental states from distributional fit.

\section{Case analysis: autonomous learning and captured evaluation}

\citet{dupoux2026} propose an architecture intended to transcend the dominant AI paradigm's limitations. Because the proposal correctly diagnoses the field's structural problem, the lossiness that reappears across its design stack and evaluation is especially diagnostic.

The proposal argues current AI lacks autonomous learning because current learning remains externalized through human-managed pipelines, and it responds by proposing lessons from cognitive science, namely an A/B/M architecture intended to integrate observation, action, and meta-control. System A handles unsupervised observation-based learning, System B handles action-based reinforcement learning, and System M provides meta-control over when and how each mode is engaged. The following analysis unpacks the proposal's operative theoretical commitments to show how they shape both its evaluation criteria and its viable engineering directions.

\subsection{Specifying the capability claim}

The claimed capability is ``autonomous learning'': the ability of a system to learn adaptively from real-world interaction without requiring human-orchestrated training pipelines. \citet{dupoux2026} define this in explicit contrast to current AI systems, which ``learn essentially nothing'' (p.~2) once deployed and whose learning ``is outsourced to human experts instead of being an intrinsic capability'' (p.~15). To overcome this limitation, the authors claim new architectures should instantiate a fundamentally different kind of learning, one that is self-directed, feedback-coupled, and adaptive across genuinely novel situations.

\subsection{Extracting theoretical assumptions}

The dominant learning theory implicit in current AI and benchmark design is distributional and performance-based. We call this the \emph{distributional performance theory}: learning is improvement on specified tasks measured against held-out data from the same distribution.

\citet{dupoux2026} invoke multiple traditions of learning beyond this theory. These traditions are not interchangeable. They offer distinct and partially incompatible accounts of what learning requires, including mechanisms the distributional framework was not designed to capture.

The authors acknowledge that the distributional account is historically contingent rather than conceptually necessary, and that learning has been studied across a ``rich set of traditions and methods in the natural, social, and formal sciences,'' but then decline to review those traditions. Specifically, cybernetic accounts \citep{wiener1948} require real-time error correction through feedback loops coupling system outputs to environmental states. Situated and sociocultural accounts \citep{lave1988,vygotsky1978} treat competence as an achievement of person-in-setting rather than a property of an isolated agent. Ecological accounts \citep{gibson1966} locate learning in active perceptual engagement with a real environment.

The paper gestures toward this broader space through ``active SSL,'' in which System B directs System A's attention across portions of a sensory stream. But this preserves the language of active perception while foreclosing its structural requirement. For Gibson, the structure available to perception is not antecedently given; it is progressively disclosed through the organism's movement through a real environment, such that the space of perceivable affordances cannot be specified in advance of the engagement that produces them. Active SSL, by contrast, operates over a sensory stream whose distribution is fixed prior to attention allocation. The available ``action'' is selection within a pre-specified input space, not the generative coupling through which ecological structure becomes perceivable in the first place. This formalization of the ecological account does not satisfy the architectural requirements specified by the theory.

More broadly, the paper's central distinction of ``autonomous learning'' (observation-based versus action-based learning) draws from AI subfield divisions. Operationally, both System A and System B are primarily supported by computational citations \citep{saffran1996,rao1999,sutton2018,schultz1997}. The operational theoretical assumptions are therefore narrower than the authors' claimed capability. This constitutes a lossy compression of the cited traditions: the mechanisms that define those accounts are reduced to forms legible within the distributional paradigm.

This resulting lossiness cannot be defended as ordinary engineering simplification. Because the capability claim is grounded in the traditions the authors themselves cite, any simplification must preserve the mechanisms those traditions treat as constitutive of learning. Here it does not: a distributional formalization of learning is not a tractable approximation of cybernetic feedback coupling; it omits the closed loop that defines cybernetic learning. A reward-optimized policy is not a tractable approximation of mediated activity; it is the formalism mediated-activity theorists developed their accounts against. This creates a critical instability: if autonomous learning requires mechanisms beyond distributional curve-fitting, substituting a reward-optimized policy undermines the claim to have instantiated it; if it does not, the proposal never departs from the paradigm it opposes.

Because the mechanism is replaced at the level of theoretical assumption, the instability propagates through the design stack, becoming visible at the level of evaluation. Autonomous learning, as defined here, is an ill-structured capability: open-ended, context-dependent, and not reducible to predefined success criteria \citep{jonassen1997}. The proposed evaluation criteria instead treat it as a well-structured proxy with fixed solution spaces and convergent performance metrics. The framework therefore converts an ill-structured capability into a well-structured form and stabilizes success on that form as evidence of the capability.

\subsection{Deriving architectural and evaluative requirements}

These theoretical differences define different units of analysis and therefore imply different requirements of architecture, training, and evaluation. The distributional account that dominates current AI systems is operationalized through loss minimization over static datasets: objective functions reward output similarity to a target distribution, and evaluation tests performance on held-out samples from that same distribution.

Autonomous learning requires more than this framework can detect. A cybernetic account requires real-time feedback pathways through which system behavior is updated by environmental response, not weights fixed at deployment. A situated account requires competence to be assessed in context, not abstracted from it. An ecological account requires active perceptual coupling with an uncontrolled environment, not performance on recorded or synthetic inputs. These are not alternative measures of the same capacity. They imply different mechanisms and therefore different evaluative conditions. Evaluation must therefore test performance under structurally different settings rather than generalization within a distribution.

The substitution emerges in the paper's treatment of action. In the A/B framework, action is intervention directed toward a specified objective, a formalization inherited from reinforcement learning. In Vygotskian terms, by contrast, action is mediated by tools, signs, and social relations that transform both actor and activity \citep{vygotsky1978,leontev1978,engestrom1987}. Learning is not policy adjustment toward reward, but participation in culturally mediated practice. An agent optimizing reward over state-action trajectories is therefore not a Vygotskian subject. The cited theoretical traditions are not operationalized at the level of architectural or evaluative mechanism.

The same gap appears in the paper's appeal to cybernetics. For \citet{wiener1948}, learning requires closed-loop error correction: system behavior modifies the environment, and environmental response modifies subsequent behavior. System B is action-based, but not cybernetically coupled in this sense. The authors are explicit that System B's policies are ``learned over the organism's lifetime via gradient descent or reinforcement'' \citep[p.~10]{dupoux2026}: optimization over stored experience, not real-time environmental updating. The paper positions itself as heir to the 1950s cybernetic tradition it invokes in its foreword, then proposes an architecture that reproduces the very absence of environmental coupling that distinguished Wiener's program from its behaviorist contemporaries.

The capability claim also implies a stronger form of meta-regulation than System M is designed to satisfy. Second-order cybernetics would require not merely switching among pre-specified learning modes, but revising the criteria by which such switching is governed \citep{vonfoerster1974,maturana1980}. The authors specify the opposite: System M's ``core routing policy is hardwired, an evolutionarily fixed transition table that dictates when to explore, when to plan, and when to act'' \citep[p.~10]{dupoux2026}. Meta-regulation in their architecture is fixed at the evolutionary timescale, not adaptive at the lifetime of the agent. Even on this stronger reading, however, the core evaluative problem remains unchanged: the proposed criteria cannot distinguish genuinely self-revising meta-control from pre-specified mode switching. Trials-to-criterion and hours of language exposure are therefore blind not only to first-order feedback coupling, but also to whether the system's regulation of learning is itself adaptive in the required sense.

The analysis here is not that reinforcement learning should be rewritten to incorporate sociocultural theory; it is that a capability claim grounded in a cited tradition cannot be operationalized through a formalism that tradition was developed to refute. More precisely: these traditions do not prescribe different methods for instantiating competence; they deny that competence is the kind of thing that can be extracted from context, generalized, and transferred, which is precisely what autonomous learning, as a field-level capability claim, requires. Taking these traditions seriously as theoretical commitments would require not a different architecture within the current paradigm, but a more radical reconception of what the capability claim means and whether it is coherent on the terms the traditions themselves supply. Each of these traditions, taken seriously, undermines the notion of a singular autonomous cognitive agent; this is the very unit of analysis the capability claim presupposes.

\subsection{Testing discriminative validity}

Operationally, discriminative validity requires contrastive conditions under which systems lacking the target mechanism can succeed while systems possessing it remain uniquely identifiable. The evaluation trap in \citet{dupoux2026} is double: the proposed criteria do not test the broader theoretical traditions the paper invokes, nor do they discriminate the narrower conception it actually operationalizes.

\citet{dupoux2026} propose two integration tests: trials required to learn a new task compared to human performance, and hours of exposure needed to acquire language at child level \citep{dupoux2018}. Both operationalize learning as performance change on specified tasks. This introduces \textbf{criterion leakage}: although the benchmarks are framed as approximating developmental learning, the criteria register output equivalence while allowing the learning process itself to drop out of evaluation.

This is the transferability assumption in operation. Because the benchmark relies on \textbf{proxy substitution}, measuring task convergence rather than the learning process the capability claim requires, performance on the benchmark is treated as sufficient warrant for the broader capability. Under optimization pressure, this produces systems that succeed under benchmark conditions while remaining behaviorally approximate with respect to the capability itself.

The unit tests reproduce the same pattern at the component level: System A is evaluated on distributional generalization, System B on reward optimization, System M on switching efficiency, each operationalized within the theoretical vocabulary of its source subfield rather than derived from the capability claim. The result is \textbf{context blindness}. The cybernetic requirement for real-time feedback disruption, the sociocultural requirement for mediated task structure, and the ecological requirement for uncontrolled environment coupling each go untested because the evaluation cannot vary the conditions required to detect them.

Taken together, these failures reveal an implicit functionalism in the evaluation criteria: the assumption that if a system produces outputs indistinguishable from those of a genuinely capable system, it instantiates the capability. This extends the transferability assumption to the level of mechanism: functional equivalence is taken to license inference to mechanism equivalence, even when mechanism equivalence is precisely what the capability claim requires. The traditions \citet{dupoux2026} invoke were each developed in explicit rejection of functionalist output equivalence as the criterion of competence. Evaluating claims grounded in those traditions through functionalist criteria reproduces the reduction those traditions were constructed to refute.

This is the circularity problem made concrete. The benchmark does not merely fail to detect the capability; it reorganizes what counts as the capability in terms of its own evaluative criteria. The result is a structural ceiling: systems optimized within the benchmark regime converge on what the evaluation can detect, while the capability the benchmark claims to measure remains out of reach but no longer visible as absent. Because evaluation criteria determine what counts as successful instantiation, they also determine which architectures can function as viable research directions.

\subsection{Epistematics applied: derivation table}

Table~\ref{tab:audit} summarizes the Epistematics audit: the theoretical accounts invoked by the capability claim, the architectural requirements they imply, the evaluative mismatches produced by the proposed criteria, and the conditions required for discriminative validity.

\begin{table}[p]
\caption{Epistematics audit of \citet{dupoux2026}. Each row applies the procedure to a theoretical source invoked by the autonomous learning claim, deriving the architectural requirement, identifying the evaluation criterion used, predicting the resulting false positive, and specifying the condition required for discriminative validity.}
\label{tab:audit}
\small
\setlength{\tabcolsep}{4pt}
\begin{tabularx}{\textwidth}{>{\RaggedRight}p{2.4cm} >{\RaggedRight}p{2.6cm} >{\RaggedRight}p{2.2cm} >{\RaggedRight}p{3.0cm} >{\RaggedRight\arraybackslash}p{3.0cm}}
\toprule
\textbf{Theoretical Source} & \textbf{Architectural Requirement} & \textbf{Evaluation Criterion} & \textbf{Predicted False Positive} & \textbf{Discriminative Condition} \\
\midrule
Cybernetics \citep{wiener1948} & Real-time error-correction via environmental feedback & Trials-to-criterion & Feedback-driven adaptation indistinguishable from distributional convergence & Disrupt feedback pathways; measure adaptation change \\
\addlinespace
Situated cognition \citep{lave1988} & Context-sensitive competence in person-in-setting & Performance vs.\ human baseline & Context-sensitive competence indistinguishable from task-specific optimization & Vary task, environment, and social context unpredictably from training \\
\addlinespace
Sociocultural theory \citep{vygotsky1978} & Mediated activity through tools, signs, social relations & Hours of language exposure & Internalized cultural practice indistinguishable from statistical regularities & Require mediated tasks; measure performance contingent on mediation \\
\addlinespace
Ecological psychology \citep{gibson1966} & Active perceptual coupling with real environment & Exposure-based metrics & Organism-environment coupling indistinguishable from passive data consumption & Require real-time perception-action coupling with uncontrolled environment \\
\addlinespace
Meta-control \citep{dupoux2026} & Autonomous data selection and mode switching & Human-comparison thresholds & Internally generated switching indistinguishable from externally orchestrated & Compare against externally controlled baselines \\
\bottomrule
\end{tabularx}
\end{table}

\section{Design criteria for autonomous learning evaluation}

The required discriminative conditions identified in Table~\ref{tab:audit} generate the benchmark design criteria that follow. The claim is not that autonomous learning cannot be benchmarked, but that valid benchmarks must be derived from the requirements of the capability rather than inherited performance proxies. At minimum, evaluation must satisfy the following conditions:

\textbf{Discriminate online adaptation from pretraining effects.} Evaluation must include conditions unreachable through prior distributional optimization, with performance improvement under those conditions treated as necessary evidence of genuine online adaptation.

\textbf{Verify feedback-loop structure.} Evaluation must include interventions that selectively disrupt feedback pathways; sensitivity to disruption is necessary evidence of feedback-coupled learning.

\textbf{Test context-sensitivity across genuine context change.} Evaluation must include transfer conditions in which task, environment, and social context vary in ways not predictable from training, and stable performance across such variation is necessary to distinguish context-sensitive competence from task-specific optimization.

\textbf{Assess self-directed data selection and mode switching.} Evaluation must compare system behavior against externally controlled baselines, treating divergence from baseline behavior as evidence that data selection and mode transitions are internally generated.

\textbf{Probe the boundary of the design envelope.} Evaluation must include conditions genuinely novel relative to training rather than held-out from the same distribution; failure under such conditions, alongside success within-distribution, is diagnostic of behavioral approximation rather than general adaptive capacity.

\subsection{Open-world evaluation}

Open-world interactive environments (e.g., large-scale sandbox video games) provide one possible substrate for the evaluative conditions required by autonomous learning. These environments are formally specified causal systems with dynamic states, open-ended task structures, emergent context, and consequences that are immediate and interpretable. They also permit human performance baselines to be established across a range of tasks and contexts.

Their value lies in context sensitivity and the capacity for contrastive testing. For example, feedback pathways can be selectively disrupted to distinguish adaptive coupling from pretraining effects. Environmental affordances and task contexts can be altered to separate context-sensitive competence from task-specific optimization. Mediated task structures can be varied to determine whether competence depends on the forms of mediation invoked by the capability claim. Baselines can be constructed against pretraining-only systems, externally controlled systems, and systems optimized directly on the environment to determine whether data selection and mode switching are internally generated.

However, open-world environments can themselves become proxies if their task structures, scoring systems, and training distributions stabilize as the evaluation target. The environment must therefore function as an experimental substrate for contrastive mechanism testing, not as a leaderboard for aggregate performance.

\section{Scope and limitations}

Epistematics audits capability-evaluation coherence; it does not replace empirical benchmark validation. Its validity depends on the alignment between evaluative criteria and the theoretical requirements of the capability being assessed. This alignment presupposes capability claims specified at the level of mechanism rather than behavior, and theoretical traditions that yield derivable architectural requirements. The procedure produces candidate criteria, not validated benchmarks; empirical work determines whether the derived criteria behave as predicted. Disagreement in translating theoretical accounts into design constraints is diagnostic of competing interpretations.

\section{Conclusion}

Benchmark design and its accompanying theoretical commitments structure AI research by determining what can be recognized as progress, what can be pursued as a viable research direction, and what structural limitations remain invisible. The case of autonomous learning illustrates this mechanism: a proposal that attempts to depart from the dominant paradigm at the architectural level is reabsorbed at the level of evaluation, which imposes the assumptions that define success.

Epistematics addresses this by making the derivation from capability claim to evaluation criterion explicit, reproducible, and auditable. The procedure does not demand abandoning existing benchmarks or importing external normative standards. It requires asking whether the criteria used to assess a claimed capability are coherent with what that capability implies, and constructing evaluation frameworks that answer that question.

The failure mode taxonomy and design criteria produced here are specific to autonomous learning. The procedure that generates them is general. Applied systematically across AI capability claims (e.g., reasoning, world-modeling), Epistematics produces a diagnostic method for determining whether benchmark success means what it claims to mean, and if not, why.

\bibliographystyle{plainnat}
\bibliography{references}

\end{document}